\documentclass[11pt,a4paper]{article}
\usepackage[nohyperref]{acl2017}
\usepackage{times}
\usepackage{latexsym}
\usepackage{amsmath}
\usepackage{amsfonts}
\usepackage{amssymb}
\usepackage{url}
\usepackage{float}
\usepackage{graphicx}
\usepackage{color}

\usepackage{algorithm,algpseudocode}

\aclfinalcopy 

\newcommand{\seqtoseq}{sequence-to-sequence}
\newcommand{\Seqtoseq}{Sequence-to-sequence}
\newcommand{\smallgiga}{\textsc{Giga}-200k}
\newcommand{\mediumgiga}{\textsc{Giga}-2M}
\newcommand{\largegiga}{\textsc{Giga}-20M}

\DeclareMathOperator*{\argmaxB}{argmax}   

\title{Neural AMR: Sequence-to-Sequence Models for Parsing and Generation}

\author{Ioannis Konstas$^{\dagger}$ \quad Srinivasan Iyer$^{\dagger}$ \quad Mark Yatskar$^{\dagger}$ \quad\\ \textbf{Yejin Choi}$^{\dagger}$ \quad \textbf{Luke Zettlemoyer}$^{\dagger}$$^{\ddagger}$ \\\\
        $^{\dagger}$Paul G. Allen School of Computer Science \& Engineering, Univ. of Washington, Seattle, WA \\
         \texttt{\{ikonstas,sviyer,my89,yejin,lsz\}@cs.washington.edu}\\
        \\
         $^{\ddagger}$Allen Institute for Artificial Intelligence, Seattle, WA\\
        \texttt{lukez@allenai.org}}

\date{}

\begin{document}
\maketitle
\begin{abstract}


Sequence-to-sequence 
models have shown strong performance across a broad range of applications. However, their application to parsing and generating text using Abstract Meaning Representation (AMR) 
has been limited, 
due to the relatively limited amount of labeled data and the non-sequential nature of the AMR graphs. 
We present a novel training procedure that can lift this limitation using millions of unlabeled sentences and careful preprocessing of the AMR graphs. 
For AMR parsing, our model achieves competitive results of 62.1 SMATCH, the current best score reported without significant use of external 
semantic resources. For AMR generation, our model establishes a new state-of-the-art performance of BLEU 33.8. We present extensive ablative and qualitative analysis including 
strong evidence that sequence-based AMR models are robust   
against ordering variations of graph-to-sequence conversions. 



\end{abstract}

\section{Introduction}
\begin{figure}
    \centering
    \includegraphics[scale=.36]{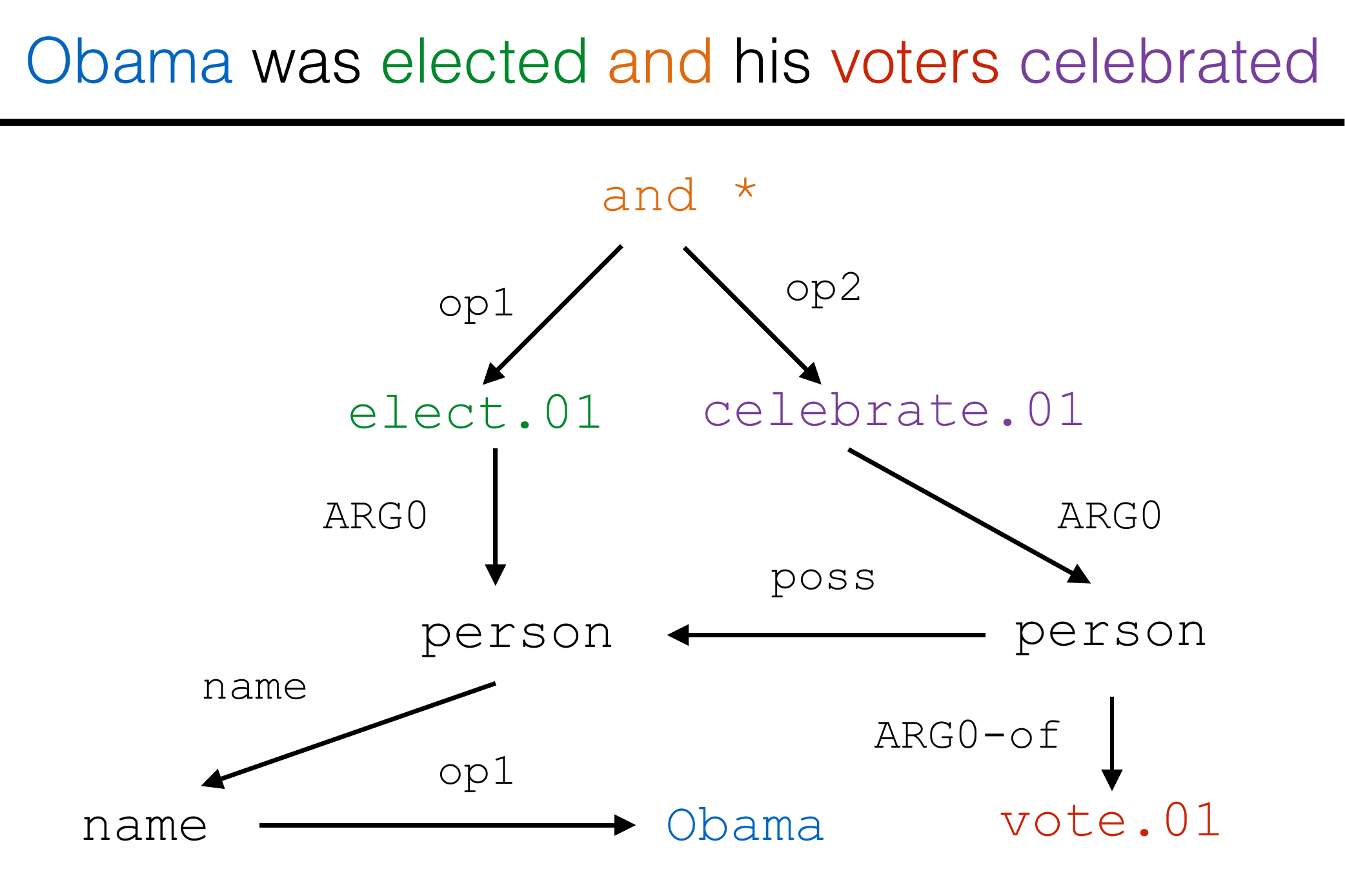}
    \caption{An example sentence and its corresponding Abstract Meaning Representation (AMR). AMR encodes semantic dependencies between entities mentioned in the sentence, such as ``Obama'' being the ``arg0'' of the verb ``elected''.}
    \label{fig:amr_graph}
\end{figure}

Abstract Meaning Representation (AMR) is 
a semantic formalism to encode the meaning of natural language text. As shown in Figure~\ref{fig:amr_graph}, AMR represents the meaning using a directed graph while abstracting away the surface forms in text. 
AMR has been used as an intermediate meaning representation for several applications including machine translation (MT) \cite{jones-EtAl:2012:PAPERS}, summarization \cite{liu-EtAl:2015:NAACL-HLT3}, sentence compression \cite{takase-EtAl:2016:EMNLP2016}, and event extraction \cite{huang2016liberal}. 
While AMR allows for rich semantic representation, annotating training data in AMR is expensive, which in turn limits the use of neural network models 
\cite{misra-artzi:2016:EMNLP2016,peng:2017:EACL,barzdins-gosko:2016:SemEval}. 

In this work, 
we present the first successful sequence-to-sequence (seq2seq) models that achieve strong results for both text-to-AMR parsing and AMR-to-text generation. 
Seq2seq models have been broadly successful in many other applications
\cite{wu2016google,bahdanau2014neural,luong-pham-manning:2015:EMNLP,NIPS2015_5635}. However, their application to AMR has been limited, 
in part because effective {\it linearization} (encoding graphs as linear sequences) and data sparsity were thought to pose significant challenges.
We show that these challenges can be easily overcome, by demonstrating that seq2seq models can be trained using {\it any} graph-isomorphic linearization and that unlabeled text can be used to significantly reduce sparsity.  


Our approach is two-fold. 
First, we introduce a novel paired training procedure that enhances both the text-to-AMR parser and AMR-to-text generator. 
More concretely, first we use self-training to bootstrap a high quality AMR parser from millions of unlabeled Gigaword sentences \cite{napoles-gormley-vandurme:2012:AKBC-WEKEX} and then use the automatically parsed AMR graphs to pre-train an AMR generator. 
This paired training allows both the parser and generator to learn high quality representations of 
fluent English text 
from millions of weakly labeled examples, that are then fine-tuned using human annotated AMR data.

Second, we propose a preprocessing procedure for the AMR graphs, which includes anonymizing entities and dates, grouping entity categories, and encoding nesting information in concise ways, as illustrated in Figure \ref{fig:preprocessing}(d). 
This preprocessing procedure helps overcoming the data sparsity while also substantially reducing the complexity of the AMR graphs.  
Under such a representation, we show that any depth first traversal of the AMR is 
an effective linearization, 
and it is even possible to use a different random order for each example.


Experiments on the LDC2015E86 AMR corpus (SemEval-2016 Task 8) demonstrate the effectiveness of the overall approach. 
For parsing, we are able to obtain competitive performance of 62.1 SMATCH without using any external annotated examples other than the output of a NER system, an improvement of over 10 points relative to neural models with a comparable setup. 
For generation, we  
substantially outperform previous best results, establishing a new state of the art of 33.8 BLEU. 
We also provide extensive ablative and qualitative analysis, quantifying the contributions that come from preprocessing and the paired training procedure.





\section{Related Work}

\paragraph{Alignment-based Parsing}
\newcite{flanigan-EtAl:2014:P14-1} (JAMR) pipeline concept and relation identification with a graph-based algorithm. 
\newcite{zhou-EtAl:2016:EMNLP20163} extend JAMR by performing the concept and relation identification tasks jointly with an incremental model. 
Both systems rely on features based on a set of alignments produced using bi-lexical cues and hand-written rules.
In contrast, our models train directly on parallel corpora, and make only minimal use of alignments to anonymize named entities.
\paragraph{Grammar-based Parsing}
\newcite{wang-EtAl:2016:SemEval} (CAMR) perform a series of shift-reduce transformations on the output of an externally-trained dependency parser, similar to~\newcite{damonte-cohen-satta:2017:EACLlong}, \newcite{brandt-EtAl:2016:SemEval}, \newcite{puzikov-kawahara-kurohashi:2016:SemEval}, and \newcite{goodman-vlachos-naradowsky:2016:SemEval}. 
\newcite{artzi-lee-zettlemoyer:2015:EMNLP} use a grammar induction approach with Combinatory Categorical Grammar (CCG), 
which relies on pre-trained CCGBank categories, like \newcite{bjerva-bos-haagsma:2016:SemEval}. 
\newcite{pust-EtAl:2015:EMNLP} recast parsing as a string-to-tree Machine Translation problem, using unsupervised alignments \cite{pourdamghani-EtAl:2014:EMNLP2014}, 
and employing several external semantic resources. 
Our neural approach is engineering lean, relying only on a large unannotated corpus of English and algorithms to find and canonicalize named entities. 
\paragraph{Neural Parsing}  Recently there have been a few seq2seq systems for AMR parsing 
 \cite{barzdins-gosko:2016:SemEval, peng:2017:EACL}. 
Similar to our approach, \newcite{peng:2017:EACL} deal with sparsity by anonymizing named entities and typing low frequency words, resulting in a very compact vocabulary (2k tokens). However, we avoid reducing our vocabulary by introducing a large set of unlabeled sentences from an external corpus, therefore drastically lowering the out-of-vocabulary rate (see Section~\ref{sec:sparsity}).
\paragraph{AMR Generation}
\newcite{flanigan-EtAl:2016:N16-1} specify a number of tree-to-string transduction rules based on alignments and POS-based features that are used to drive a tree-based SMT system. 
\newcite{pourdamghani-knight-hermjakob:2016:INLG} also use an MT decoder; they learn a classifier that linearizes the input AMR graph in an order that follows the output sentence, effectively reducing the number of alignment crossings of the phrase-based decoder.
\newcite{song-EtAl:2016:EMNLP2016} recast generation as a traveling salesman problem, after partitioning the graph into fragments and finding the best linearization order. 
Our models do not need to rely on a particular linearization of the input, attaining comparable performance even with a per example random traversal of the graph. 
Finally, all three systems intersect with a large language model trained on Gigaword. 
We show that our seq2seq model has the capacity to learn the same information as a language model, especially after pretraining on the external corpus.
\paragraph{Data Augmentation} Our paired training procedure is largely inspired by \newcite{sennrich-haddow-birch:2016:P16-11}. They improve neural MT performance for low resource language pairs by using a back-translation MT system for a large monolingual corpus of the target language in order to create synthetic output, and mixing it with the human translations. We instead pre-train on the external corpus first, and then fine-tune on the original dataset. 


\section{Methods}
In this section, we first provide the formal definition of AMR parsing and generation (section~\ref{ssec:task}). Then we describe the \seqtoseq~models we use (section~\ref{ssec:s2s}), graph-to-sequence conversion (section~\ref{sec:linearization_setup}), and our paired training procedure 
(section~\ref{ssec:training}). 

\subsection{Tasks}
\label{ssec:task}
We assume access to a training dataset $D$ where each example pairs a natural language sentence $s$ with an AMR  $a$.
The AMR is a rooted directed acylical graph.
It contains nodes whose names correspond to sense-identified verbs, nouns, or AMR specific concepts, for example \texttt{elect.01}, \texttt{Obama}, and \texttt{person}  in Figure~\ref{fig:amr_graph}.
One of these nodes is a distinguished root, for example, the node \texttt{and} in Figure~\ref{fig:amr_graph}.  
Furthermore, the graph contains labeled edges, which correspond to PropBank-style ~\cite{Palmer-propbank} semantic roles for verbs or other relations introduced for AMR, for example, \texttt{arg0} or \texttt{op1} in Figure~\ref{fig:amr_graph}.
The set of node and edge names in an AMR graph is drawn from a set of tokens $C$, and every word in a sentence is drawn from a vocabulary $W$. 

We study the task of training an {\bf AMR parser}, i.e., finding a set of parameters $\theta_P$ for model $f$, that predicts an AMR graph $\hat{a}$, given a sentence $s$:

\begin{equation}\label{eq:parsing}
\hat{a} = \argmaxB_{a} f\bigl( a | s; \theta_P \bigr)
\end{equation}

We also consider the reverse task, training an {\bf AMR generator} by finding a set of parameters $\theta_G$, for a model $f$ that predicts a sentence $\hat{s}$, given an AMR graph $a$:

\begin{equation}\label{eq:generation}
\hat{s} = \argmaxB_{s} f\bigl( s | a; \theta_G \bigr)
\end{equation}

In both cases, we use the same family of predictors $f$, sequence-to-sequence models that use global attention, but the models have independent parameters, $\theta_P$ and $\theta_G$. 

\subsection{\Seqtoseq~Model}
\label{ssec:s2s}
For both tasks, we use a stacked-LSTM \seqtoseq~neural architecture employed in neural machine translation~\cite{bahdanau2014neural,wu2016google}.\footnote{We extended the Harvard NLP seq2seq framework from \text{\url{http://nlp.seas.harvard.edu/code}}.}
Our model uses a global attention decoder and unknown word replacement with small modifications  \cite{luong-pham-manning:2015:EMNLP}.

The model uses a stacked bidirectional-LSTM encoder to encode an input sequence and a stacked LSTM to decode from the hidden states produced by the encoder.
We make two modifications to the encoder: (1) we concatenate the forward and backward hidden states at every level of the stack instead of at the top of the stack, and (2) introduce dropout in the first layer of the encoder. 
The decoder predicts an attention vector over the encoder hidden states using previous decoder states.
The attention is used to weigh the hidden states of the encoder and then predict a token in the output sequence.
The weighted hidden states, the decoded token, and an attention signal from the previous time step (input feeding) are then fed together as input to the next decoder state.  
The decoder can optionally choose to output an unknown word symbol, in which case the predicted attention is used to copy a token directly from the input sequence into the output sequence. 

\subsection{Linearization}
\label{sec:linearization_setup}
Our seq2seq models require that both the input and target be presented as a linear sequence of tokens.
We define a linearization order for an AMR graph as any sequence of its nodes and edges. 
A linearization is defined as (1) a linearization order and (2) a rendering function that generates any number of tokens when applied to an element in the linearization order (see Section~\ref{sec:preprocessing_linearization} for implementation details). 
Furthermore, for parsing, a valid AMR graph must be recoverable from the linearization.


\renewcommand{\algorithmicrequire}{\textbf{Input:}}
\renewcommand{\algorithmicensure}{\textbf{Output:}}
\algnewcommand{\LeftComment}[1]{\Statex \(\triangleright\) #1}
\begin{figure}
\begin{tabular}{@{}p{7cm}}
\vspace{-0.235in}
\begin{algorithm}[H]
\caption{Paired Training Procedure}
\begin{algorithmic}[1]
\footnotesize
\Require Training set of sentences and AMR graphs $(s,a) \in \mathcal{D}$, an unannotated external corpus of sentences $S_e$, a number of self training iterations, $N$, and an initial sample size $k$.
\Ensure Model parameters for AMR parser $\theta_P$ and AMR generator $\theta_G$.
\State $\theta_P \gets$ Train parser on D
\LeftComment{Self-train AMR parser.}
\State $S_{e}^{1} \gets$ sample $k$ sentences from $S_e$
\For{$i=1$ to $N$}\label{alg:main:preprocess}\label{alg:learn:ckyloop}
\State $A_e^{i} \gets $ Parse $S_e^{i}$ using parameters $\theta_P$
\LeftComment{Pre-train AMR parser.}
\State $\theta_P \gets$ Train parser on $(A_e^{i}, S_e^{i})$
\LeftComment{Fine tune AMR parser.}
\State $\theta_P \gets$ Train parser on D with initial parameters $\theta_P$
\State $S_e^{i+1} \gets$ sample $k \cdot 10^{i}$ new sentences from $S_e$
\EndFor
\State $S_e^{N} \gets$ sample $k \cdot 10^{N}$ new sentences from $S_e$
\LeftComment{Pre-train AMR generator.}
\State $A_e \gets $ Parse $S_e^N$ using parameters $\theta_P$
\State $\theta_G \gets$ Train generator on $(A_e^N,S_e^N)$
\LeftComment{Fine tune AMR generator.}
\State $\theta_G \gets$  Train generator on $D$ using initial parameters $\theta_G$
\State return $\theta_P$, $\theta_G$ 
\end{algorithmic}
\label{alg:data_augmentation} 
\end{algorithm}
\vspace{-0.6in}
\end{tabular}
\end{figure}

\subsection{Paired Training}
\label{ssec:training}
Obtaining a corpus of jointly annotated pairs of sentences and AMR graphs is expensive and current datasets only extend to thousands of examples. 
Neural sequence-to-sequence models suffer from sparsity with so few training pairs.
To reduce the effect of sparsity, we use an external unannotated corpus of sentences $S_e$, and a procedure which pairs the training of the parser and generator.

Our procedure is described in Algorithm \ref{alg:data_augmentation}, and first trains a parser on the dataset $D$ of pairs of sentences and AMR graphs.
Then it uses self-training to improve the initial parser.
Every iteration of self-training has three phases: (1) parsing samples from a large, unlabeled corpus $S_e$, (2) creating a new set of parameters by training on $S_e$, and (3) fine-tuning those parameters on the original paired data. 
After each iteration, we increase the size of the sample from $S_e$ by an order of magnitude. 
After we have the best parser from self-training, we use it to label AMRs for $S_e$ and pre-train the generator.
The final step of the procedure fine-tunes the generator on the original dataset $D$.



\section{AMR Preprocessing}\label{sec:preprocessing}

\begin{figure*}
    \centering
    \includegraphics[scale=.28]{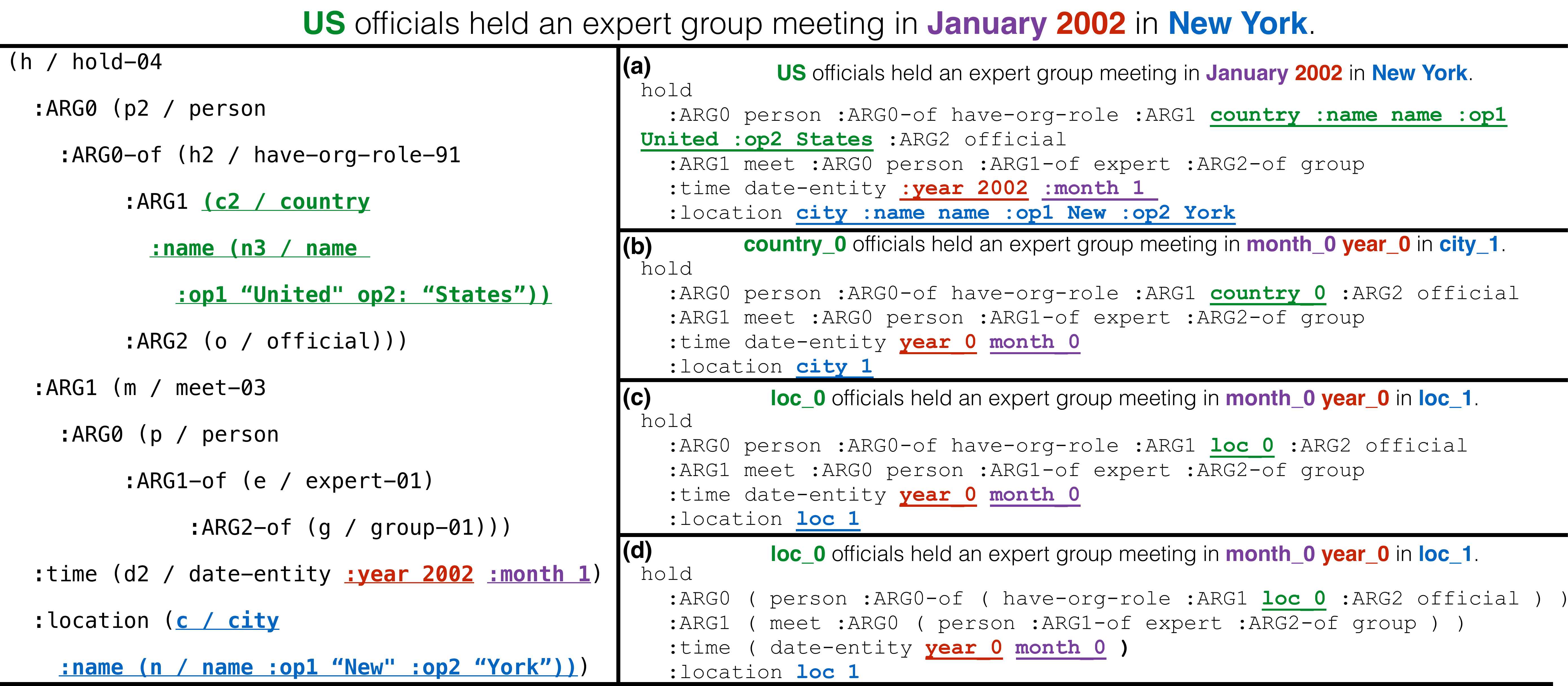}
    \caption{Preprocessing methods applied to sentence (top row) - AMR graph (left column) pairs. Sentence-graph pairs after (a) graph simplification, (b) named entity anonymization, (c) named entity clustering, and (d) insertion of scope markers.}
    \label{fig:preprocessing}
\end{figure*}


We use a series of preprocessing steps, including AMR linerization, anonymization, and other modifications we make to sentence-graph pairs.
Our methods have two goals: (1) reduce 
the complexity of the linearized sequences 
to make learning easier while maintaining enough original information, and (2) address sparsity from certain open class vocabulary entries, such as named entities (NEs) and quantities.
Figure~\ref{fig:preprocessing}(d) contains example inputs and outputs with all of our preprocessing techniques.  

\paragraph{Graph Simplification} In order to reduce the overall length of the linearized graph, we first remove variable names and the \texttt{instance-of} relation ( \texttt{/} ) before every concept. 
In case of re-entrant nodes we replace the variable mention with its co-referring concept. 
Even though this replacement incurs loss of information, often the surrounding context helps recover the correct realization, e.g., the possessive role \texttt{:poss} in the example of Figure~\ref{fig:amr_graph} is strongly correlated with the surface form \textit{his}.
Following \newcite{pourdamghani-knight-hermjakob:2016:INLG} we also remove senses from all concepts for AMR generation only.
Figure~\ref{fig:preprocessing}(a) contains an example output after this stage.

\subsection{Anonymization of Named Entities}
Open-class types including NEs, dates, and numbers account for 9.6\% of tokens in the sentences of the training corpus, and 31.2\% of vocabulary $W$.
83.4\% of them occur fewer than 5 times in the dataset.
In order to reduce sparsity and be able to account for new unseen entities, we perform extensive anonymization.

First, we anonymize sub-graphs headed by one of AMR's over 140 fine-grained entity types that contain a \texttt{:name} role. 
This captures structures referring to entities such as \texttt{person}, \texttt{country}, miscellaneous entities marked with \texttt{*-enitity}, and typed numerical values, \texttt{*-quantity}. 
We exclude \texttt{date} entities (see the next section).
We then replace these sub-graphs with a token indicating fine-grained type and an index, $i$, indicating it is the $i$th occurrence of that type.\footnote{In practice we only used three groups of ids: a different one for NEs, dates and constants/numbers.}
For example, in Figure~\ref{fig:preprocessing} the sub-graph headed by \texttt{country} gets replaced with \texttt{country$\_$0}. 

On the training set, we use alignments obtained using the JAMR aligner \cite{flanigan-EtAl:2014:P14-1} and the unsupervised aligner of \newcite{pourdamghani-EtAl:2014:EMNLP2014} in order to find mappings of anonymized subgraphs to spans of text and replace mapped text with the anonymized token that we inserted into the AMR graph.
We record this mapping for use during testing of generation models.
If a generation model predicts an anonymization token, we find the corresponding token in the AMR graph and replace the model's output with the most frequent mapping observed during training for the entity name.
If the entity was never observed, we copy its name directly from the AMR graph.

\paragraph{Anonymizing Dates} For dates in AMR graphs, we use separate anonymization tokens for year, month-number, month-name, day-number and day-name, indicating whether the date is mentioned by word or by number.\footnote{We also use three date format markers that appear in the text as: \textit{YYYYMMDD}, \textit{YYMMDD}, and \textit{YYYY-MM-DD}.}
In AMR generation, we render the corresponding format when predicted.
Figure~\ref{fig:preprocessing}(b) contains an example of all preprocessing up to this stage.

\paragraph{Named Entity Clusters}
When performing AMR generation, each of the AMR fine-grained entity types is manually mapped to one of the four coarse entity types used in the Stanford NER system~\cite{Finkel:2005:INI:1219840.1219885}: person, location, organization and misc.
This reduces the sparsity associated with many rarely occurring entity types.
Figure~\ref{fig:preprocessing} (c) contains an example with named entity clusters.
\paragraph{NER for Parsing}
When parsing, we must normalize test sentences to match our anonymized training data.
To produce fine-grained named entities, we run the Stanford NER system and first try to replace any identified span with a fine-grained category based on alignments observed during training.
If this fails, we anonymize the sentence using the coarse categories predicted by the NER system, which are also categories in AMR. 
After parsing, we deterministically generate AMR for anonymizations using the corresponding text span.

\subsection{Linearization}\label{sec:preprocessing_linearization}
\label{sec:preprocessing_lin}

\paragraph{Linearization Order} Our linearization order is defined by the order of nodes visited by depth first search, including backward traversing steps.
For example, in Figure~\ref{fig:preprocessing}, starting at \texttt{meet} the order contains \texttt{meet}, \texttt{:ARG0}, \texttt{person}, \texttt{:ARG1-of}, \texttt{expert}, \texttt{:ARG2-of}, \texttt{group}, \texttt{:ARG2-of}, \texttt{:ARG1-of},  \texttt{:ARG0}.\footnote{Sense, \texttt{instance-of} and variable information has been removed at the point of linearization.}
The order traverses children in the sequence they are presented in the AMR.
We consider alternative orderings of children in Section~\ref{sec:linearization} but always follow the pattern demonstrated above.

\paragraph{Rendering Function} Our rendering function marks scope, and generates tokens following the pre-order traversal of the graph: 
(1) if the element is a node, it emits the type of the node. 
(2) if the element is an edge, 
it emits the type of the edge and then recursively emits a bracketed string for the (concept) node immediately after it. In case the node has only one child we omit the scope markers (denoted with left ``\texttt{(}'', and right ``\texttt{)}'' parentheses), thus significantly reducing the number of generated tokens.
Figure~\ref{fig:preprocessing}(d) contains an example showing all of the preprocessing techniques and scope markers that we use in our full model.

\begin{table*}
    \centering    
    \begin{tabular}{l|ccc||ccc}      
              & \multicolumn{3}{|c||}{Dev} & \multicolumn{3}{|c}{Test} \\\hline
        Model & Prec & Rec & F1 & Prec & Rec & F1\\\hline        
        {\sc SBMT} \cite{pust-EtAl:2015:EMNLP}  & - & - & 69.0 & - & - & 67.1 \\
        {\sc JAMR} \cite{S16-1186} & - & - & - & 69.7 & 64.5 & 67.0 \\
        {\sc CAMR} \cite{wang-EtAl:2016:SemEval} & 72.3 & 61.4 & 66.6 & 70.4 & 63.1 & 66.5 \\
        {\sc CCG}* \cite{artzi-lee-zettlemoyer:2015:EMNLP} & 67.2 & 65.1 & 66.1 & 66.8 & 65.7 & 66.3 \\
        {\sc JAMR} \cite{flanigan-EtAl:2014:P14-1} & - & - & - & 64.0 & 53.0 & 58.0 \\\hline  
        \largegiga & 62.2 &  66.0 & 64.4 & 59.7 & 64.7 & 62.1 \\   
        \mediumgiga & 61.9 & 64.8 & 63.3 & 60.2 & 63.6 & 61.9 \\   
        \smallgiga & 59.7 & 62.9 & 61.3 & 57.8 & 60.9 & 59.3\\
        {\sc AMR-only} & 54.9 & 60.0 & 57.4 & 53.1 & 58.1 & 55.5 \\\hline     
        {\sc seq2seq} \cite{peng:2017:EACL} & - & - & - & 55.0 & 50.0 & 52.0 \\
        {\sc char-lstm} \cite{barzdins-gosko:2016:SemEval} & - & - & - & - & - & 43.0 \\
    \end{tabular}
    \caption{SMATCH scores for AMR Parsing. *Reported numbers are on the newswire portion of a previous release of the corpus (LDC2014T12).}
    \label{tab:pretraining_parsing}    
\end{table*}

\section{Experimental Setup}

We conduct all experiments on the AMR corpus used in SemEval-2016 Task 8 (LDC2015E86), which contains 16,833/1,368/1,371 train/dev/test examples.
For the paired training procedure of Algorithm~\ref{alg:data_augmentation}, we use Gigaword as our external corpus and sample sentences that only contain words from the AMR corpus vocabulary $W$. We sub-sampled the original sentence to ensure there is no overlap with the AMR training or test sets. Table~\ref{tab:dataset_stats} summarizes statistics about the original dataset and the extracted portions of Gigaword. 
We evaluate AMR parsing with SMATCH \cite{cai-knight:2013:Short}, and AMR generation using BLEU \cite{papineni-EtAl:2002:ACL}\footnote{We use the multi-BLEU script from the MOSES decoder suite \cite{Koehn:2007:MOS:1557769.1557821}.}.

We validated word embedding sizes and RNN hidden representation sizes by maximizing AMR development set performance (Algorithm~\ref{alg:data_augmentation} -- line 1). 
We searched over the set \{128, 256, 500, 1024\} for the best combinations of sizes and set both to 500.
Models were trained by optimizing cross-entropy loss with stochastic gradient descent, using a batch size of 100 and dropout rate of 0.5.
Across all models when performance does not improve on the AMR dev set, we decay the learning rate by 0.8. 

For the initial parser trained on the AMR corpus, (Algorithm~\ref{alg:data_augmentation} -- line 1), we use a single stack version of our model, set initial learning rate to 0.5 and train for 60 epochs, taking the best performing model on the development set. 
All subsequent models benefited from increased depth and we used 2-layer stacked versions, maintaining the same embedding sizes.
We set the initial Gigaword sample size to $k=200,000$ and executed a maximum of 3 iterations of self-training. 
For pre-training the parser and generator, (Algorithm~\ref{alg:data_augmentation} -- lines 4 and 9), we used an initial learning rate of 1.0, and ran for 20 epochs.
We attempt to fine-tune the parser and generator, respectively, after every epoch of pre-training, setting the initial learning rate to 0.1.
We select the best performing model on the development set among all of these fine-tuning attempts.
During prediction we perform decoding using beam search and set the beam size to 5 both for parsing and generation.

\section{Results}

\begin{table}
    \centering
    
    \begin{tabular}{l|ccc}
        Corpus & Examples & OOV@1 & OOV@5 \\\hline
        AMR & 16833 & 44.7 & 74.9 \\
        \smallgiga & 200k & 17.5 & 35.3 \\
        \mediumgiga & 2M & 11.2 & 19.1 \\
        \largegiga & 20M & 8.0 & 12.7\\
    \end{tabular}
    
    \caption{LDC2015E86 AMR training set, \smallgiga, \mediumgiga~and \largegiga~statistics; OOV@1 and OOV@5 are the out-of-vocabulary rates on the NL side with thresholds of 1 and 5, respectively. Vocabulary sizes are 13027 tokens for the AMR side, and 17319 tokens for the NL side.}
    \label{tab:dataset_stats}
\end{table}

\begin{table}[t]
    \centering
    \begin{tabular}{@{~}l@{~}|@{~}c@{~}||@{~}c@{~}}        
        Model & Dev & Test\\\hline        
        \largegiga & 33.1 & 33.8 \\
        \mediumgiga & 31.8 & 32.3\\
        \smallgiga & 27.2 & 27.4 \\
        {\sc AMR-only} & 21.7 & 22.0 \\\hline
        {\sc PBMT}* \cite{pourdamghani-knight-hermjakob:2016:INLG} & 27.2 & 26.9 \\
        {\sc TSP} \cite{song-EtAl:2016:EMNLP2016} & 21.1 & 22.4 \\
        {\sc TreeToStr} \cite{flanigan-EtAl:2016:N16-1} & 23.0 & 23.0 \\        
    \end{tabular}
    \caption{BLEU results for AMR Generation. *Model has been trained on a previous release of the corpus (LDC2014T12).}
    \label{tab:pretraining_generation}    
\end{table}

\paragraph{Parsing Results}
Table~\ref{tab:pretraining_parsing} summarizes our development results for different rounds of self-training and test results for our final system, self-trained on 200k, 2M and 20M unlabeled Gigaword sentences.
Through every round of self-training, our parser improves. 
Our final parser outperforms comparable seq2seq and character LSTM models by over 10 points. 
While much of this improvement comes from self-training, our model without Gigaword data outperforms these approaches by 3.5 points on F1. We attribute this increase in performance to different handling of preprocessing and more careful hyper-parameter tuning.
All other models that we compare against use semantic resources, such as WordNet, dependency parsers or CCG parsers (models marked with * were trained with less data, but only evaluate on newswire text; the rest evaluate on the full test set, containing text from blogs).
Our full models outperform the original version of JAMR \cite{flanigan-EtAl:2014:P14-1}, a graph-based model but still lags behind other parser-dependent systems (CAMR\footnote{Since we are  currently not using any Wikipedia resources for the prediction of named entities, we compare against the no-wikification version of the CAMR system.}), and resource heavy approaches (SBMT).
\paragraph{Generation Results}
Table~\ref{tab:pretraining_generation} summarizes our AMR generation results on the development and test set. 
We outperform all previous state-of-the-art systems by the first round of self-training and further improve with the next rounds.
Our final model trained on \largegiga~outperforms TSP and {\sc TreeToStr} trained on LDC2015E86, by over 9 BLEU points.\footnote{We also trained our generator on \mediumgiga~and fine-tuned on LDC2014T12 in order to have a direct comparison with PBMT, and achieved a BLEU score of 29.7, i.e., 2.8 points of improvement.} 
Overall, our model incorporates less data than previous approaches as all reported methods train language models on the whole Gigaword corpus.
We leave scaling our models to all of Gigaword for future work.

\paragraph{Sparsity Reduction}\label{sec:sparsity}
Even after anonymization of open class vocabulary entries, we still encounter a great deal of sparsity in vocabulary given the small size of the AMR corpus, as shown in Table~\ref{tab:dataset_stats}.
By incorporating sentences from Gigaword we are able to reduce vocabulary sparsity dramatically, as we increase the size of sampled sentences: the out-of-vocabulary rate with a threshold of 5 reduces almost 5 times for~\largegiga.



\begin{table}
    \centering
    \begin{tabular}{l|c}
        Model & BLEU\\\hline
        {\sc Full} & 21.8\\
        {\sc Full - scope} & 19.7\\
        {\sc Full - scope - ne} & 19.5\\
        {\sc Full - scope - ne - anon} & 18.7 \\
    \end{tabular}
    \caption{BLEU scores for AMR generation ablations on preprocessing (DEV set).}
    \label{tab:preprocessing_generation}
\end{table}
\begin{table}
    \centering
    \begin{tabular}{l|ccc}
        Model & Prec & Rec & F1 \\\hline
        {\sc Full} & 54.9 & 60.0 & 57.4\\
        {\sc Full - anon} & 22.7 & 54.2 & 32.0 \\
    \end{tabular}
    \caption{SMATCH scores for AMR parsing ablations on preprocessing (DEV set).}
    \label{tab:preprocessing_parsing}
\end{table}

\paragraph{Preprocessing Ablation Study}
We consider the contribution of each main component of our preprocessing stages while keeping our linearization order identical.
Figure~\ref{fig:preprocessing} contains examples for each setting of the ablations we evaluate on. 
First we evaluate using linearized graphs without parentheses for indicating scope, Figure~\ref{fig:preprocessing}(c), 
then without named entity clusters, Figure~\ref{fig:preprocessing}(b), and additionally without any anonymization, Figure~\ref{fig:preprocessing}(a).  

Tables~\ref{tab:preprocessing_generation} summarizes our evaluation on the AMR generation.
Each components is required, and scope markers and anonymization contribute the most to overall performance. 
We suspect without scope markers our seq2seq models are not as effective at capturing long range semantic relationships between elements of the AMR graph.
We also evaluated the contribution of anonymization to AMR parsing (Table~\ref{tab:preprocessing_parsing}). 
Following previous work, we find that seq2seq-based AMR parsing is largely ineffective without anonymization~\cite{peng:2017:EACL}.



\section{Linearization Evaluation}
\label{sec:linearization}

\begin{table}[t]
    \centering
    \begin{tabular}{l|c}
        Linearization Order & BLEU  
        \\\hline
        {\sc Human} & 21.7 \\
        {\sc Global-Random} & 20.8\\
        {\sc Random} & 20.3 \\
    \end{tabular}
    \caption{BLEU scores for AMR generation for different linearization orders (DEV set).}
    \label{tab:linearization}
\end{table}

In this section we evaluate three strategies for converting AMR graphs into sequences in the context of AMR generation and show that our models are largely agnostic to linearization orders. 
Our results argue, unlike SMT-based AMR generation methods~\cite{pourdamghani-knight-hermjakob:2016:INLG}, that seq2seq models can learn to ignore artifacts of the conversion of graphs to linear sequences.


\subsection{Linearization Orders}
All linearizations we consider use the pattern described in Section~\ref{sec:preprocessing_lin}, but differ on the order in which children are visited. 
Each linearization generates anonymized, scope-marked output (see Section~\ref{sec:preprocessing}), of the form shown in Figure~\ref{fig:preprocessing}(d). 

\paragraph{Human}
The proposal traverses children in the order presented by human authored AMR annotations exactly as shown in Figure~\ref{fig:preprocessing}(d).

\paragraph{Global-Random}
We construct a random global ordering of all edge types appearing in AMR graphs and re-use it for every example in the dataset. 
We traverse children based on the position in the global ordering of the edge leading to a child.

\paragraph{Random} For each example in the dataset we traverse children following a different random order of edge types.

\subsection{Results}
We present AMR generation results for the three proposed linearization orders in Table~\ref{tab:linearization}.
Random linearization order performs somewhat worse than traversing the graph according to Human linearization order. 
Surprisingly, a per example random linearization order performs nearly identically to a global random order, arguing seq2seq models can learn to ignore artifacts of the conversion of graphs to linear sequences.

\paragraph{Human-authored AMR leaks information}
The small difference between random and global-random linearizations argues that our models are largely agnostic to variation in linearization order.
On the other hand, the model that follows the human order performs better, which leads us to suspect it carries extra information not apparent in the graphical structure of the AMR. 

To further investigate, we compared the relative ordering of edge pairs under the same parent to the relative position of children nodes derived from those edges in a sentence, as reported by JAMR alignments.
We found that the majority of pairs of AMR edges (57.6\%) always occurred in the same relative order, therefore revealing no extra generation order information.\footnote{This is consistent with constraints encoded in the annotation tool used to collect AMR. For example, \texttt{:ARG0} edges are always ordered before \texttt{:ARG1} edges.}
Of the examples corresponding to edge pairs that showed variation, 70.3\% appeared in an order consistent with the order they were realized in the sentence. 
The relative ordering of some pairs of AMR edges was particularly indicative of generation order.
For example, the relative ordering of edges with types \texttt{location} and \texttt{time}, was 17\% more indicative of the generation order than the majority of generated locations before \texttt{time}.\footnote{Consider the sentences \textit{``She went to school in New York two years ago''}, and \textit{``Two years ago, she went to school in New York''}, where \textit{``two year ago''} is the time modifying constituent for the verb \textit{went} and \textit{``New York''} is the location modifying constituent of \textit{went}.}

To compare to previous work we still report results using human orderings. However, we note that any practical application requiring a system to generate an AMR representation with the intention to realize it later on, e.g., a dialog agent,  will need to be trained either using consistent, or random-derived linearization orders. Arguably, our models are agnostic to this choice.







 \begin{table}[t]
     \centering
     \begin{tabular}{l|c}
         Error Type  & \% \\\hline                        
         Coverage & 29 \\
         Disfluency & 23\\          
         Anonymization & 14\\         
         Sparsity & 13\\ 
         Attachment & 12 \\ 
         Other & 10\\        
    \end{tabular}
    \caption{Error analysis for AMR generation on a sample of 50 examples from the development set.}
   \label{tab:errors}
 \end{table}
 
\section{Qualitative Results}

\begin{figure}
    \centering
    \includegraphics[scale=.36]{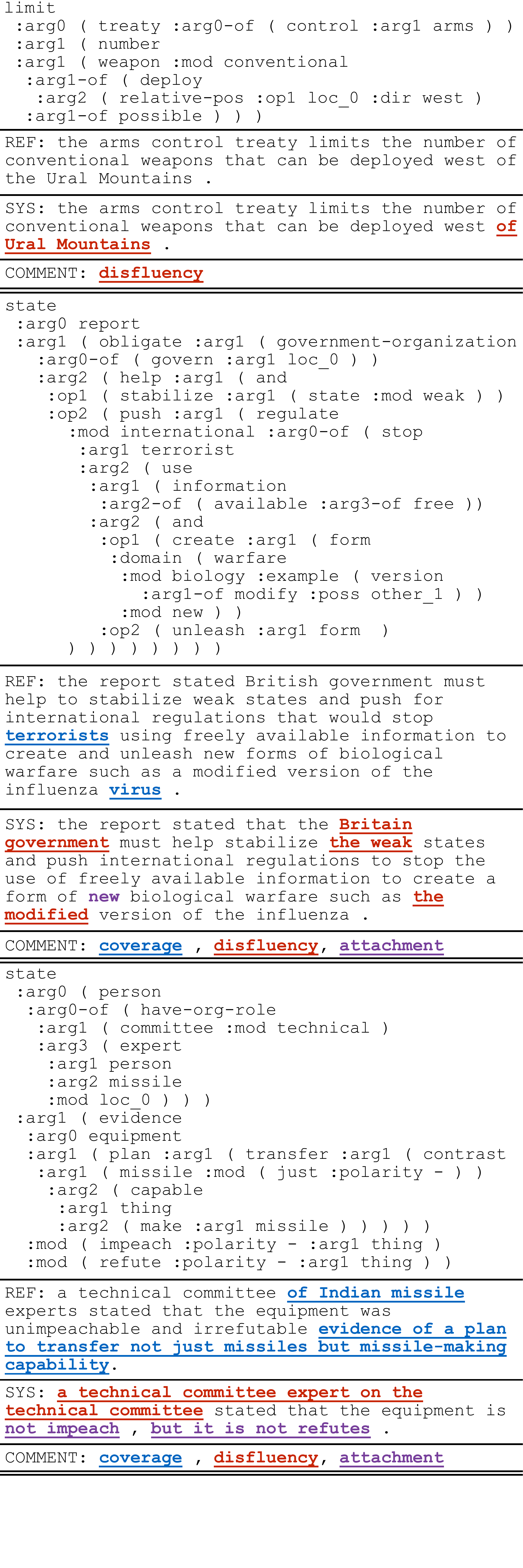}
    \vspace{-2cm}
    \caption{Linearized AMR after preprocessing, reference sentence, and output of the generator.
    We mark with colors common error types: disfluency, coverage (missing information from the input graph), and attachment (implying a semantic relation from the AMR between incorrect entities).
    }    
    \label{fig:qual}
\end{figure}

Figure~\ref{fig:qual} shows example outputs of our full system. 
The generated text for the first graph is nearly perfect with only a small grammatical error due to anonymization.
The second example is more challenging, with a deep right-branching structure, and a coordination of the verbs \texttt{stabilize} and \texttt{push} in the subordinate clause headed by \texttt{state}.
The model omits some information from the graph, namely the concepts \texttt{terrorist} and \texttt{virus}.
In the third example there are greater parts of the graph that are missing, such as the whole sub-graph headed by \texttt{expert}. 
Also the model makes wrong attachment decisions in the last two sub-graphs (it is the \texttt{evidence} that is \textit{unimpeachable} and \textit{irrefutable}, and not the \texttt{equipment}), mostly due to insufficient annotation (\texttt{thing}) thus making their generation harder.

Finally, Table~\ref{tab:errors} summarizes the proportions of error types we identified on 50 randomly selected examples from the development set. We found that the generator mostly suffers from coverage issues, an inability to mention all tokens in the input, followed by fluency mistakes, as illustrated above. Attachment errors are less frequent, which supports our claim that the model is robust to graph linearization, and can successfully encode long range dependency information between concepts.

\section{Conclusions}

We applied sequence-to-sequence models to the tasks of AMR parsing and AMR generation, by carefully preprocessing the graph representation and scaling our models via pretraining on millions of unlabeled sentences sourced from Gigaword corpus. Crucially, we avoid relying on resources such as knowledge bases and externally trained parsers. We achieve competitive results for the parsing task (SMATCH 62.1) and state-of-the-art performance for generation (BLEU 33.8). 

For future work, we would like to extend our work to different meaning representations such as the Minimal Recursion Semantics (MRS; \newcite{Copestake2005}). This formalism tackles certain linguistic phenomena differently from AMR (e.g., negation, and co-reference), contains explicit annotation on concepts for number, tense and case, and finally handles multiple languages\footnote{A list of actively maintained languages can be found here: \url{http://moin.delph-in.net/GrammarCatalogue}} \cite{BENDER14.639.L14-1508}. Taking a step further, we would like to apply our models on Semantics-Based Machine Translation using MRS as an intermediate representation between pairs of languages, and investigate the added benefit compared to directly translating the surface strings, especially in the case of distant language pairs such as English and Japanese \cite{Siegel:Bender:Bond:2016}.

\section*{Acknowledgments}
\small
The research was supported in part by DARPA under the DEFT program through AFRL (FA8750-13-2-0019) and the CwC program through
 ARO (W911NF-15-1-0543), the ARO (W911NF-16-1-0121), the NSF (IIS-1252835, IIS-1562364, IIS-1524371), an Allen Distinguished Investigator Award, Samsung GRO, and gifts by Google
and Facebook. The authors thank Rik Koncel-Kedziorski, the UW NLP group,
and the anonymous reviewers for their thorough and helpful comments.

{
\bibliography{acl2017}
\bibliographystyle{acl_natbib}
}
\end{document}